\documentclass[conference]{IEEEtran}
\IEEEoverridecommandlockouts
\usepackage{cite}
\usepackage{algorithmic}
\usepackage[linesnumbered,ruled]{algorithm2e}
\usepackage{amsmath,amssymb,amsfonts}
\usepackage{algorithmic}
\usepackage{graphicx}
\usepackage{subcaption}
\graphicspath{{Figures/}}
\usepackage{textcomp}
\usepackage{xcolor}
\def\BibTeX{{\rm B\kern-.05em{\sc i\kern-.025em b}\kern-.08em
    T\kern-.1667em\lower.7ex\hbox{E}\kern-.125emX}}

\begin{document}

\title{Revisiting semi-supervised training objectives for differentiable particle filters\\
}

\author{\IEEEauthorblockN{Jiaxi Li,
John-Joseph Brady,
Xiongjie Chen, 
and
Yunpeng Li}
\IEEEauthorblockA{Computer Science Research Centre,
University of Surrey, Guildford, United Kingdom}
\IEEEauthorblockA{Emails: \{jiaxi.li, j.brady, xiongjie.chen, yunpeng.li\}@surrey.ac.uk}
}

\maketitle

\begin{abstract}
Differentiable particle filters combine the flexibility of neural networks with the probabilistic nature of sequential Monte Carlo methods. However, traditional approaches rely on the availability of labelled data, i.e., the ground truth latent state information, which is often difficult to obtain in real-world applications. This paper compares the effectiveness of two semi-supervised training objectives for differentiable particle filters. We present results in two simulated environments where labelled data are scarce.
\end{abstract}


\section{Introduction}

Particle filters are a class of sequential state estimation algorithms with utility in fields such as robotics, signal processing, and target tracking~\cite{gordon1993novel,doucet2000sequential,dai2019visual, palmier2019adaptive, qian20173d}. However, classical particle filtering requires prior knowledge of the functional form of the underlying state space model (SSM)~\cite{doucet2001introduction, kantas2015particle, coskun2017long}. Differentiable particle filters (DPFs) address this limitation by parameterising the SSM with neural networks~\cite{jonschkowski18, karkus2018particle}. DPFs are trained by stochastic gradient descent, requiring back-propagation through the filtering algorithm~\cite{corenflos2021differentiable}. A review of differentiable particle filtering methods and components can be found in~\cite{chen2023overview}.

Directly optimising the accuracy of the filtering mean requires access to the ground truth latent state during training, however in real world scenarios this is typically not readily available. While generating precise ground truth latent state can be difficult, obtaining observations is relatively cheap.  Most models of practical interest, including normalising flow DPFs~\cite{chen2022conditional} and the models considered in visual robot localisation~\cite{jonschkowski18, karkus2018particle}, are non-identifiable given the observations alone, hereafter referred to as the `unsupervised setting'. We hypothesise that augmenting conventional supervised learning over a limited dataset with unsupervised learning techniques over a larger dataset can lead to improved performance. Learning procedures where ground truth is available only over a limited portion of the training dataset are referred to as `semi-supervised'~\cite{berthelot2019remixmatch}.

In~\cite{wen2021end}, a semi-supervised training objective combining a supervised loss and an unsupervised pseudo-likelihood~\cite{andrieu2005} based loss was used. The experimental results presented in~\cite{wen2021end} appear to justify its usage; however, no ablation study was conducted to verify whether the performance improvement is due to the proposed training loss or other modification e.g. the updated measurement model in~\cite{wen2021end}. We attempt to verify the performance of pseudo-likelihood loss in this work.
Furthermore, there has been no exploration of semi-supervised training objectives based on the, more popular, evidence lower-bound (ELBO)~\cite{le2018auto,maddison2017filtering,naesseth2018variational,li2023} unsupervised loss.

The contribution of this paper is primarily experimental: in a pair of simulated-data experiments, we demonstrate the efficacy of incorporating the ELBO loss in a semi-supervised training objective function over the pseudo-likelihood~\cite{wen2021end}. This paper serves to provide guidance for practitioners in training differentiable particle filters with limited data.

The rest of this article is structured as follows: Section~\ref{sec:ps} presents the problem statement. We provide preliminaries on differentiable particle filtering in Section~\ref{sec:preliminaries}. Section~\ref{sec:objectives} describes the two considered semi-supervised training objectives. Section~\ref{sec:results} presents the experimental setup and results. We conclude the paper in Section~\ref{sec:conclusion}.

\section{Problem Statement}
\label{sec:ps}
In this section, we address a nonlinear filtering task characterised by the following SSM:
\begin{align}
	\label{eq:state_spaces_model}
	&x_0 \sim \pi\left(x_0\right)\,\,, 
	\\ \label{dyn} 
	&x_t \sim p_{\theta}\left(x_t \mid x_{t-1}\right)\,\,,
	\\ \label{mea} 
	&y_t \sim p_{\theta}\left(y_t \mid x_t \right)\,\,,
\end{align}
where the initial hidden state follows the prior distribution $\pi(x_0)$, and the dynamic model $p_\theta(x_t|x_{t-1})$ describes the evolution of hidden states. The measurement model $p_\theta(y_t|x_t)$ relates the observation to the hidden state at the same time step. $\theta$ denotes the parameters of the SSM.

Our objective is to learn the parameter set $\theta$ and perform sequential posterior inference. To improve sampling efficiency, we use a proposal distribution $q_{\theta}(x_{t}| y_t, x_{t-1})$ that uses information from the latest observation to sample particles. We focus on solving this Bayesian filtering task in a setting where a limited number of data points are labelled with their ground truth latent state.

\section{Preliminaries}
\label{sec:preliminaries}
\subsection{Particle filtering}
\label{subsec:pf}
Particle filters approximate the latent state distribution at time $t$, given observations up to that time, by a set of importance weighted particles:

\begin{equation}
    p(x_{t}| y_{1:t};\theta) \approx \sum_{i=1}^{N_p} \mathbf{w}_t^i \delta_{x_{t}^i}(x_{t}),
\end{equation}
where $N_p$ is the number of particles, $\delta_{x_{t}^i}(\cdot)$ is a Dirac delta function centred at particle $x_{t}^i$. $\mathbf{w}_t^i$ is the normalised weight of the $i$-th particle such that $\sum_{i=1}^{N_p} \mathbf{w}_t^i = 1$.

In filtering scenarios, particles are first sampled from their prior distribution and, for subsequent times, from the proposal distribution. The case where the proposal distribution is chosen to be equal to the dynamic model is known as the `bootstrap particle filter'~\cite{gordon1993novel}. The particle weights are updated based on the likelihood of the observed data given the particle states. To mitigate the issue of weight degeneracy, particles are optionally resampled based on their weights~\cite{douc2005comparison,elvira2022rethinking}. A full tutorial on particle filtering can be found in~\cite{doucet2009tutorial}. Note that DPFs predominantly involve batch processing during training, the filtering phase still entails sequential inference~\cite{li2023,mastrototaro2024onlinevariational}. Despite the availability of complete data trajectories, which would typically warrant the application of smoothing techniques, we adhere to filtering approaches in this work due to the constraints inherent in the current framework of DPFs.

\subsection{Normalising flow-based particle filters}
\label{subsec:nf-dpf}
Normalising flow-based differentiable particle filters (NF-DPFs)~\cite{chen2021differentiable, chen2022conditional} leverage a family of tractable distributions and invertible mappings to model the system dynamics and proposal distributions with no prior knowledge of the SSM's functional form. The components of the NF-DPFs are constructed as follows.
\subsubsection{Dynamic model with normalising flows}
The dynamic model is specified using a base distribution $g_{\theta}(\cdot | x_{t-1})$ and an invertible map $\mathcal{T}_\theta(\cdot)$~\cite{chen2021differentiable}, such that:
\begin{equation}
    x_t = \mathcal{T}_\theta(\dot{x}_t), \quad \dot{x}_t \sim g_{\theta}(\dot{x}_t | x_{t-1}),
\end{equation}
with the probability density of $x_t$ derived by the change of variable formula using the Jacobian determinant of $\mathcal{T}_\theta(\cdot)$.

\subsubsection{Proposal distribution with conditional normalising flows}
A conditional normalising flow, $\mathcal{F}_\theta(\cdot)$, is used to transport particles from the dynamic model to the proposal distribution~\cite{chen2021differentiable}:

\begin{equation}
    x^i_t = \mathcal{F}_\theta\left(\mathcal{T}_\theta(\dot{x}^{i}_t); y_t\right), \quad \dot{x}^i_t \sim g_{\theta}(\dot{x}^i_t | x^i_{t-1}),
\end{equation}
again applying the change of variable formula for computing the proposal density.
\subsubsection{Measurement model with conditional normalising flows}
The measurement model is constructed using another conditional normalising flow $\mathcal{G}_\theta(\cdot)$:

\begin{equation}
    y_t = \mathcal{G}_\theta(z_t; x_t),
\end{equation}
with density derived similarly to the other normalising flows used~\cite{chen2022conditional}.

\section{Revisiting the unsupervised and semi-supervised
training objectives}
\label{sec:objectives}

In this section, we introduce the training objectives considered in this paper. We classify the base training objectives into two categories, supervised losses and unsupervised losses indicating their use of ground truth latent state, or not, respectively. Semi-supervised losses are constructed as a linear combination of a supervised loss and an unsupervised loss. We provide pseudocode for the semi-supervised DPF (SDPF) in Algorithm~\ref{alg:semi_general}.

\begin{algorithm}[htbp]
    \begin{algorithmic}[1]
    \caption{Semi-supervised differentiable particle filters. All operations indexed by $i$ should be repeated for all $i \in \{1, \dots, N_p\}$.}
    \label{alg:semi_general}
    
    \REQUIRE
        \begin{footnotesize}	
            prior distribution $\pi\left(x_0\right)$
            \hspace{2em}  marginal likelihood $p_\theta(y_{t}|x_{t})$\\
            \hspace{1.5em} dynamic model $p_\theta(x_{t}|x_{t-1})$
            \hspace{0em} proposal $q_\theta(x_{t}|x_{t-1})$\\
            \hspace{1.5em} learning rates $\alpha_{t}$  
            \hspace{5em} number of particles $N_p$\\
            \hspace{1.5em} trajectory length $T$ \\
        \end{footnotesize}
        \vspace{0.3em}
    \ENSURE
        \begin{footnotesize}
            optimised model parameters $\theta$\\
        \end{footnotesize}
        \vspace{0.5em}
        
    \STATE \textbf{Initialise} $\theta$ randomly;
    \WHILE{$\theta$ not converged}
    \STATE Draw particles $x_0^i \sim \pi\left(x_0\right)$; 
    \STATE Set importance weights $w^i_0 = p_{\theta}\left(y_t | x^i_0\right)$
    \FOR{$t=1$ to $T$}
    % \STATE Sample $A_t^i\sim\textbf{Mult}(\textbf{w}_t^1,\cdots,\textbf{w}_t^{N_p}) \; \forall i \in \{1,\cdots,N_{p}\}$
     
    \IF{Resampling condition met}
        \STATE Set the resampled indices and weights, $A^{i}_{t}, \tilde{w}^{i}_{t-1}$ with soft-resampling~\cite{karkus2018particle};
    \ELSE
        \STATE $A^{i}_{t} = i, \; \tilde{w}^{i}_{t-1} = \textbf{w}^{i}_{t-1}$;
    \ENDIF
    \STATE  $x^{i}_{t} \sim q_{\theta}\left(x_t^i | x_{t-1}^{A_t^i}, y_t\right)$;
    \STATE  $w_t^i= \tilde{w}_{t-1}^i\dfrac{p_{\theta}\left(x_t^i | x_{t-1}^{A_t^i}\right) p_{\theta}\left(y_t | x_t^i\right)}{q_{\theta}\left(x_t^i | x_{t-1}^{A_t^i}, y_t \right)}$;
    \STATE $\textbf{w}_t^i=\frac{w_t^i}{\sum_{i=1}^{N_p}w_{t}^i}$;
    \vspace{3pt}
    
%           \STATE Compute the effective sample size: $\mathrm{ESS}_t=\frac{1}{\sum_{i=1}^{N_p}\left(\textbf{w}_t^{i}\right)^2}$;
    % \IF{$\text{ESS}_t<\text{ESS}_\text{thres}$}
    % \STATE Sample $A_t^i\sim\textbf{Mult}(\textbf{w}_t^1,\cdots,\textbf{w}_t^{N_p})$ for $\forall i$ to obtain $\left\{\tilde{w}_{t}^i=1\right\}_{i=1}^{N_p}, \left\{\tilde{x}_t^{i} = x_t^{A_t^i}\right\}_{i=1}^{N_p}$;
%           \ELSE 
%           \STATE $\left\{\tilde{w}_{t}^i={w}_{t}^i\right\}_{i=1}^{N_p}$, $\left\{\tilde{x}_t^{i} = x_t^{i}\right\}_{i=1}^{N_p}$;
    % \ENDIF
        \ENDFOR

    \STATE Compute $\mathcal{L}(\theta)$ using Eq.~\eqref{eq:semi_elbo} or~\eqref{eq:semi_pl};
        \STATE $\theta \leftarrow \theta-\alpha_{t} \nabla_\theta \mathcal{L}(\theta)$;
    \ENDWHILE
	\end{algorithmic}
\end{algorithm}

\subsection{Supervised loss}
The supervised loss we used in this paper is the mean squared error (MSE) between true states and estimated states defined as follows:
\begin{gather}
\label{eq:mse}
    \mathcal{L}_{\text{MSE}}(\theta) = \frac{1}{T} \sum_{t=1}^{T} \left\|x_{t}^* - \hat{x}_{t}\right\|_2^2,
\end{gather}
where $x^*_t$ denotes the true states and $\hat{x}_t$ represents the estimated states.

\subsection{Unsupervised loss}
\subsubsection{Evidence lower bound}
The first unsupervised loss we consider is a variational inference inspired evidence lower bound (ELBO)~\cite{le2018auto,maddison2017filtering,naesseth2018variational}. We define the ELBO as the logarithm of an unbiased particle estimate of the likelihood~\cite{delmoral2014particleintro,chopin2020particle}:
\begin{gather}
\label{eq:smc_estimator}
	\mathcal{L}_{\text{ELBO}}(\theta)=\log\;\hat{p}_{\theta}(y_{1: T})=\sum_{t=1}^{T}\log\left[\frac{1}{N_{p}}\sum_{i=1}^{N_p} {w}_t^i\right]\,,
\end{gather}
where $w^{i}_{t}$ are the unnormalised weights, as defined in Alg. \ref{alg:semi_general}.
It follows from Jensen's inequality that the expectation of $\mathcal{L}_{\text{ELBO}}$ is a lower bound to the evidence:
\begin{align}
	&\log\;{p}_{\theta}(y_{1: T})  = \log\;\mathbb{E}\Big[\hat{p}_{\theta}(y_{1: T})\Big] \label{eq:eq1} \\
	 &\geq\mathbb{E}\Big[\log\;\hat{p}_{\theta}(y_{1: T})\Big] = \mathbb{E}\Big[\mathcal{L}_{\text{ELBO}}(\theta)\Big] \label{eq:eq2}\,,
\end{align}
where the expectation is taken over all random variables in the filtering algorithm.

Using the $\mathcal{L}_{\text{ELBO}}(\theta)$ to optimise the proposal is theoretically justified for smoothing trajectories in~\cite{le2018auto,naesseth2018variational}. It has also been successfully applied to filtering tasks in~\cite{mastrototaro2024onlinevariational, cox2024gaussianmixture}, and~\cite{corenflos2021differentiable} where it was used in combination with $\mathcal{L}_{\text{MSE}}(\theta)$ in a fully supervised setting.

\subsubsection{Pseudo likelihood}
The second unsupervised loss we consider is the pseudo-likelihood (PL) loss, based on~\cite{andrieu2005}, and adapted for a DPF context in~\cite{wen2021end}. 

Through dividing the observations and states into $m$ blocks of length $\mathcal{L}$, this unsupervised training objective can be expressed as:
\begin{equation}
	\mathcal{L}_{\text{PL}}(\theta) =\frac{1}{m} \sum_{b=0}^{m-1} \hat{Q}\left(\theta, \theta_b\right),
\end{equation}
where $\hat{Q}\left(\theta, \theta_b\right)$ is an estimate of the pseudo-likelihood at the \textit{b}-th block. Specifically, it can be computed as:
\begin{equation}
	\begin{aligned}
		\hat{Q}\left(\theta, \theta_b\right) 
		& =\sum_{i=1}^{N_p} w_b^i \log \big[\mu_\theta\left(x_{b L+1}^i\right) p_\theta\left(y_{b L+1} \mid x_{b L+1}^i\right)\big. \\
		& \big.\prod_{m=b L+2}^{(b+1) L} p_\theta\left(x_m^i \mid x_{m-1}^i\right) p_\theta\left(y_m \mid x_m^i\right)\big],
	\end{aligned}
\end{equation}
where $w_b^i$ is the particle weight of the $i^{\text{th}}$ particle at block $b$. This procedure assumes the existence of a stationary distribution of the Markov dynamic process, $\mu_\theta\left(x_{t}\right)$, which we assume to be uniform over some sub-set of latent space with finite Lebesgue measure.

\subsection{Semi-supervised loss}
With the above training objectives, two semi-supervised training objectives for DPFs can be defined as follows:
\begin{gather}
    \mathcal{L}_{\text{SDPF-ELBO}}(\theta)=\mathcal{L}_{\text{MSE}}(\theta) + \lambda_1 * \mathcal{L}_{\text{ELBO}}(\theta) \,\,,\label{eq:semi_elbo}\\
    \mathcal{L}_{\text{SDPF-PL}}(\theta)=\mathcal{L}_{\text{MSE}}(\theta) + \lambda_2 * \mathcal{L}_{\text{PL}}(\theta)\,\,,\label{eq:semi_pl}
\end{gather}
where $\lambda_1$ and $\lambda_2$ are the coefficients that scale the corresponding unsupervised loss. We select these coefficients through grid search in cross validation, and the candidate values are in the range of $10^3$ to $10^{-6}$ with steps of powers of 10.

In unsupervised settings, there is no direct information provided by the data about the latent state. The learned SSM needs to be carefully constructed with strong priors to be identifiable. This makes highly flexible formulations such as the NF-DPF introduced in Section~\ref{subsec:nf-dpf} unsuitable. The research hypothesis is that with partially labelled data we can mitigate this issue without requiring fully specified trajectories. 
% Note also, the proposal is included only for filtering efficiency, it makes little sense to try to optimise it with respect to a likelihood.

\section{Experiment results}
\label{sec:results}

In this section, we verify the performance of the two semi-supervised training objectives for differentiable particle filters in a multivariate linear Gaussian~\cite{naesseth2018variational,corenflos2021differentiable}, and a simulated maze environment~\cite{jonschkowski18,corenflos2021differentiable}. We adopt the root mean squared error (RMSE) between the true state and the predicted state as the evaluation metric. To explore the performance of the filters with different amounts of ground truth state information for training, we vary the labelled data percentage from the set $\{0.1\%,0.5\%,1\%,5\%,10\%,50\%,100\%\}$ for both setups\footnote{Code is available at: https://github.com/JiaxiLi1/Semi-Supervised-Learning-Differentiable-Particle-Filters.}.

\subsection{Multivariate linear Gaussian model} \label{multi_linear}
\subsubsection{Setup}
We first consider a multivariate linear Gaussian SSM formulated in~\cite{naesseth2018variational,corenflos2021differentiable}:
\begin{gather}
	\label{linear gaussian}
	x_t | x_{t-1} \sim \mathcal{N}\left(\tilde{\boldsymbol{\theta}}_1 x_{t-1}, \mathbf{I}_{d_{\mathcal{X}}}\right), x_0 \sim \mathcal{N}\left(\mathbf{0}_{d_{\mathcal{X}}}, \mathbf{I}_{d_{\mathcal{X}}}\right), \\
	y_t | x_t \sim \mathcal{N}\left(\tilde{\boldsymbol{\theta}}_2 x_t, 0.1\mathbf{I}_{d_{\mathcal{X}}}\right)\,,
\end{gather}
where $\mathbf{0}_{d_{\mathcal{X}}}$ is the $d_{\mathcal{X}} \times d_{\mathcal{X}}$ null matrix, $\mathbf{I}_{d_{\mathcal{X}}}$ is the $d_{\mathcal{X}} \times d_{\mathcal{X}}$ identity matrix. $\tilde{\boldsymbol{\theta}}:=\left\{\tilde{\boldsymbol{\theta}}_1\in\mathbb{R}^{d_{\mathcal{X}}\times d_{\mathcal{X}}}, \tilde{\boldsymbol{\theta}}_2\in \mathbb{R}^{d_{\mathcal{X}}\times d_{\mathcal{X}}}\right\}$ is the model parameters. Our target is to track the latent state $x_t$ given observation $y_t$. We set $d_{\mathcal{X}}$ to $10$.
We set the $(i,j)$-th entry of $\tilde{\boldsymbol{\theta}}_{1}$ as $\tilde{\boldsymbol{\theta}}_{1}(i, j)=\left(0.42^{|i-j|+1}\right)$, and $\tilde{\boldsymbol{\theta}}_{2} = 0.5\mathbf{I}_{d_{\mathcal{X}}}$. 

We simulate 50 trajectories from the true model, each containing 100 time steps. We divide the generated data into different batches for training, validation and testing, with a split ratio of 3:1:1, respectively. We initialise particles at time step $0$ uniformly over a hypercube whose boundaries are defined by the minimum and maximum values of the true states in the training data set for each dimension. We name the SDPFs using $\mathcal{L}_{\text{Semi-ELBO}}(\theta)$ and $\mathcal{L}_{\text{Semi-PL}}(\theta)$ as $\text{SDPF-ELBO}$ and $\text{SDPF-PL}$, respectively. The baseline method is the $\text{DPF}$ using $\mathcal{L}_{\text{MSE}}(\theta)$ as the training objective. Experiments results are generated with 20 sets of simulations involving randomly initialised trajectories and algorithms.

\subsubsection{Results}
\begin{figure}[htbp]
	\centering

	\begin{subfigure}[t]{0.495\linewidth}
		% \rule{\textwidth}{0.75\textwidth}
		\includegraphics[width=\textwidth]{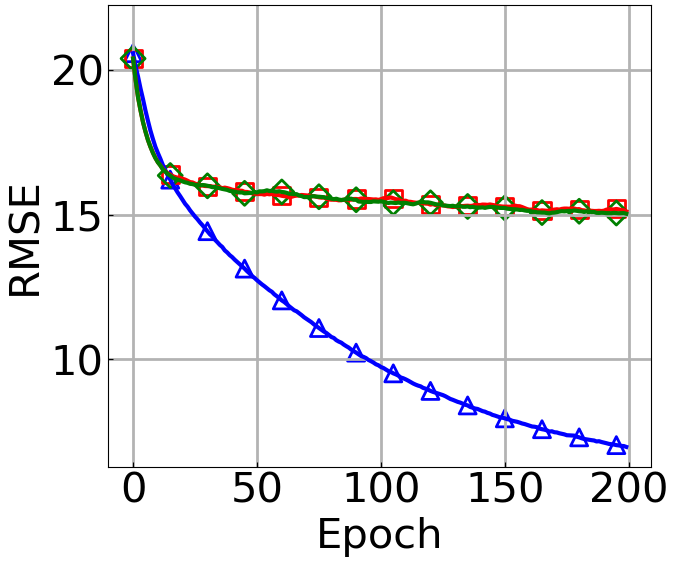}
		%\caption{Title for A}
		\caption{}
		\label{fig:linear3_a}
	\end{subfigure}\hspace{0.01pt}
	% \begin{subfigure}[t]{0.314\linewidth}
	% 	% \rule{\textwidth}{0.75\textwidth}
	% 	\includegraphics[width=\textwidth]{Figures/lin_dim10_boxplot.png}
	% 	\caption{}
	% 	\label{fig:linear3_b}
	% \end{subfigure}\hspace{0.01pt}
	\begin{subfigure}[t]{0.49\linewidth}
		% \rule{\textwidth}{0.75\textwidth}
		\includegraphics[width=\textwidth]{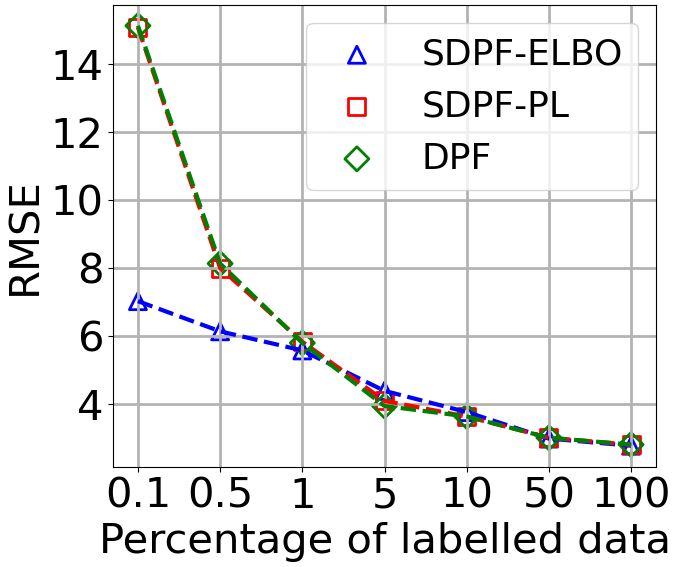}
		\caption{}
		\label{fig:linear3_c}
	\end{subfigure}\par
	\caption{The comparison of RMSEs using the DPF, the SDPF-PL, and the SDPF-ELBO for $d_{\mathcal{X}} =10$. (a) Mean RMSEs on validation data during training, with a labelling ratio of $0.1\%$. (b): Mean RMSEs on testing data with varying labelling ratios.}
	\label{fig: multivariate linear}
\end{figure}

The experimental results are illustrated in Figure \ref{fig: multivariate linear}. Figure \ref{fig:linear3_a} was produced with a labelling ratio of $0.1\%$, showing that latent state values are only available in $0.1\%$ of the time steps in the training dataset. We observe the faster convergence of the SDPF-ELBO validation RMSE at the final time-step when the labelling ratio is $0.1\%$. Figure \ref{fig:linear3_c} shows the tracking performance under different labeling ratios, indicating that the performance of the SDPF-ELBO is the best when the amount of labelled time steps is small, while the benefit of the SDPF-ELBO becomes negligible when there is a sufficient amount of labelled data, i.e. over $1\%$ in this example. 

The SDPF-PL and the DPF lead to similar performance as seen in Figure~\ref{fig: multivariate linear}.
In cross validation, we found that for different percentages of labelled data, the optimal parameter of $\lambda_2$ is always small (less than $10^{-5}$), rendering the contribution of pseudo-likelihood loss is negligible. Model performance deteriorates if we increase the value of $\lambda_2$, suggesting that the SDPF-PL is ineffective in this setting. These experiments were repeated with $d_{\mathcal{X}} \in \{2,5\}$ with similar results.

\subsection{Simulated maze environment}
\subsubsection{Experiment setup}
In this section, the proposed method is examined in a simulated maze environment~\cite{beattie2016deepmind}. This experiment relates to track the pose of the robot from a sequence of simulated view-point images and odometry data. 

The state is the coordinates for the position and orientation of the robot, which ${x}_t=\left(x_{1, t}, x_{2, t}, \omega_t\right)^{\top}$. The actions $a_{t}$ of the robot are the noisy odometry data. The observations are view-point images and are, noised and randomly cropped to $24\times24, 32\times32$ RGB images. 

The dynamic model is given as:
\begin{gather}
x_t=x_{t-1} + \mathbf{A}_t{a}_t+\xi_t,\\
\mathbf{A}_t=\begin{pmatrix}
            \cos{\omega_t} & \sin{\omega_t} & 0 \\
            \sin{\omega_t} & -\cos{\omega_t} & 0 \\
            0 & 0 & 1  \\
            \end{pmatrix},\\
\xi_t\stackrel{\text { i.i.d. }}{\sim} \mathcal{N}\left(\mathbf{0}, \operatorname{diag}\left(\sigma_x^2, \sigma_x^2, \sigma_\omega^2\right)\right),
\end{gather}
where $\xi_t$ is the dynamic noise with $\sigma_x=20$ and $\sigma_\omega=0.5$.

There are $2000$ trajectories, each with $100$ time steps. We adopt a train : validation : test split of $900:100:1000$ on the number of trajectories in each set. We initialise 100 particles uniformly across the maze. We use the Adam optimiser~\cite{kingma2015adam} to optimise DPFs and set the learning rate as 0.001.

We find that the NF-DPFs introduced in Section~\ref{subsec:nf-dpf} produced unstable results in settings where only a small proportion of data is labelled, e.g. labelled ratio = 0.1, so we follow the bootstrap particle filtering setup in~\cite{wen2021end}. For measurement models, we also follow the setup in~\cite{wen2021end}, where $p_\theta(y_t|x_t)$ is assumed to be proportional to the cosine distance between the observation and particle state encodings. 
% In this formulation the likelihood is not normalised and so it is unclear to us what the effect of optimising the 'ELBO' or 'PL' should be, although the latter is empirically justified in~\cite{wen2021end}.
In this setting, we refer to the bootstrap analogues as follows: the differentiable bootstrap particle filter (DBPF) for the DPF, and similarly, the SDBPF-ELBO and the SDBPF-PL as the bootstrap versions of the SDPF-ELBO and the SDPF-PL, respectively. We use an auto-encoder loss to ensure that the encodings retain key information of observations~\cite{wen2021end}, for comparison we include two additional baselines, the SDBPF-PL-vanilla and the DBPF-vanilla which do not use an auto-encoder loss. For each algorithm, 20 simulations were performed through random initialisation.

\subsubsection{Experiment results}

\begin{figure}[!ht]
	\centering

	\begin{subfigure}[t]{0.49\linewidth}
 % \rule{\textwidth}{0.75\textwidth}
		\includegraphics[width=\textwidth]{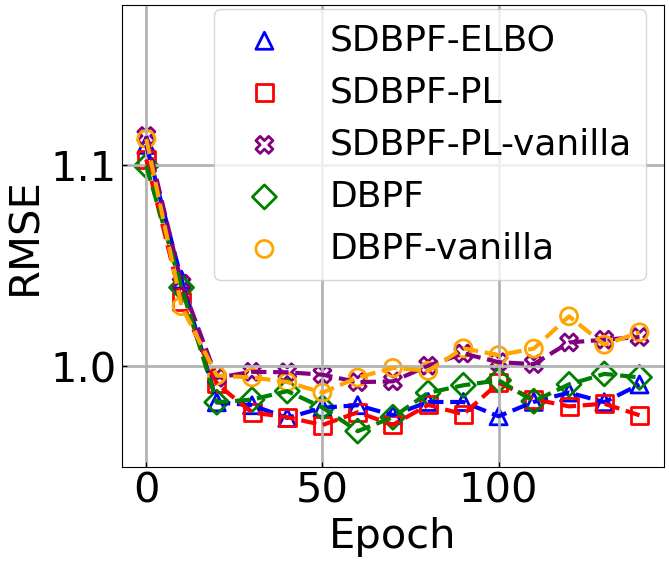}
		\caption{}
		\label{fig:maze1_a}
	\end{subfigure}\hspace{0.01pt}
	% \begin{subfigure}[t]{0.305\linewidth}
	% 	% \rule{\textwidth}{0.75\textwidth}
	% 	\includegraphics[width=\textwidth]{Figures/boxplot_maze1.png}
	% 	%\caption{}
	% 	\label{fig:maze1_b}
	% \end{subfigure}\hspace{0.01pt}
	\begin{subfigure}[t]{0.49\linewidth}
		% \rule{\textwidth}{0.75\textwidth}
		\includegraphics[width=\textwidth]{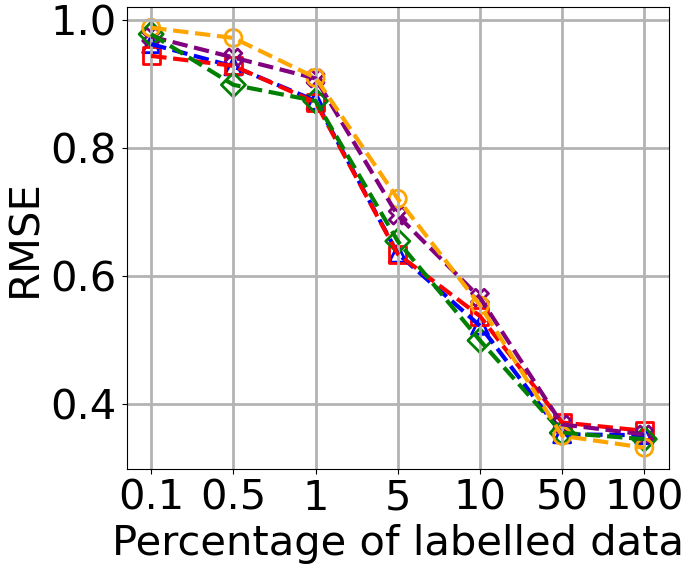}
		\caption{}
		\label{fig:maze1_d}
	\end{subfigure}\par
	
	\caption{The comparison of RMSEs between the SDBPF-ELBO, the SDBPF-PL, the SDBPF-PL-vanilla, the DBPF and the DBPF-vanilla. (a): Mean RMSEs on validation data during training, with a labelling ratio of $0.1\%$. (b): Mean RMSEs on testing data with varying labelling ratios.}
	\label{fig:maze-dbpf}
\end{figure}

In Figure \ref{fig:maze-dbpf}, the RMSEs between the SDBPF-ELBO, the SDBPF-PL, and the DBPF are similar, suggesting a marginal impact of the examined semi-supervised learning losses over traditional supervised methods.

This finding contrasts with~\cite{wen2021end} which finds performance gain of the SDBPF-PL over the DBPF in the same experiment setting. Several reasons can contribute to this, including the lack of ablation studies of other algorithmic changes introduced in \cite{wen2021end}, such as the autoencoder loss. The DBPF-vanilla, instead of the DBPF, was compared with the SDBPF-PL in \cite{wen2021end}. We show in Figure~\ref{fig:maze-dbpf} that while the estimation errors from the SDBPF-PL and the DBPF are similar, the SDBPF-PL performs marginally better than the DBPF-vanilla, showing the impact of the autoencoder loss rather than the pseudo-likelihood-based loss term adopted in \cite{wen2021end}.

%Figure~\ref{fig:maze1_d} shows that when the label data is less than 50\%, the testing RMSE of both SDBPF-PL-vanilla and DBPF-vanilla is slightly higher than that of other algorithms.

\section{Conclusion}
\label{sec:conclusion}

This paper explores the efficacy of two semi-supervised training objectives applied to differentiable particle filters, which integrate a supervised training objective with two candidate unsupervised objectives: one based on the pseudo-likelihood and the other on the ELBO of the marginal log-likelihood. We performed algorithmic comparisons in two environments: a multivariate linear model and a simulated maze environment. The results indicate that the integration of the ELBO into semi-supervised training enhances model performance in one linear Gaussian simulation setup with limited labelled data but a similar performance gain from the ELBO-based loss is not seen in a simulated maze environment. Contrary to the findings presented in \cite{wen2021end}, our findings indicate that the pseudo-likelihood loss does not enhance the performance of the SDPFs in either experimental setup, thus its use is not recommended in the contexts examined in this paper. Future research directions include more comprehensive experiment evaluation and more effective training objectives for semi-supervised training.

\clearpage
\bibliographystyle{IEEEtran}
\bibliography{ref}

\end{document}